\documentclass[letterpaper]{article} 
\usepackage{aaai24}  
\usepackage{times}  
\usepackage{helvet}  
\usepackage{courier}  
\usepackage[hyphens]{url}  
\usepackage{graphicx} 
\usepackage{natbib}  
\usepackage{caption} 
\usepackage{algorithm}
\usepackage{algorithmic}

\usepackage{enumitem}
\usepackage{multirow}
\usepackage{makecell}
\usepackage{colortbl}
\usepackage[table,xcdraw]{xcolor}
\usepackage{amssymb}
\usepackage{amsmath}
\usepackage{xspace}

\usepackage{newfloat}
\usepackage{listings}
\DeclareCaptionStyle{ruled}{labelfont=normalfont,labelsep=colon,strut=off} 
\floatstyle{ruled}
\newfloat{listing}{tb}{lst}{}
\floatname{listing}{Listing}
\title{AltNeRF: Learning Robust Neural Radiance Field\\via Alternating Depth-Pose Optimization}

\author{
Kun Wang\textsuperscript{\rm 1}, Zhiqiang Yan\textsuperscript{\rm 1}, Huang Tian\textsuperscript{\rm 1}, Zhenyu Zhang\textsuperscript{\rm 2}, Xiang Li\textsuperscript{\rm 3}, Jun Li\textsuperscript{\rm 1}\thanks{Corresponding authors} and Jian Yang\textsuperscript{\rm 1}\footnotemark[1]
}
\affiliations{
\textsuperscript{\rm 1}PCA Lab, Nanjing University of Science and Technology, China\\
\textsuperscript{\rm 2}Nanjing University, Suzhou Campus, China\\
\textsuperscript{\rm 3}Nankai University, China\\
\{kunwang, Yanzq, tianhuang, xiang.li.implus, junli, csjyang\}@njust.edu.cn, zhangjesse@foxmail.com
}
\usepackage{bibentry}

\makeatletter
\DeclareRobustCommand\onedot{\futurelet\@let@token\@onedot}
\def\@onedot{\ifx\@let@token.\else.\null\fi\xspace}
\def\eg{\emph{e.g}\onedot} 
\def\ie{\emph{i.e}\onedot}

\makeatother

\begin{document}
\maketitle
\begin{abstract}
    Neural Radiance Fields (NeRF) have shown promise in generating realistic novel views from sparse scene images. However, existing NeRF approaches often encounter challenges due to the lack of explicit 3D supervision and imprecise camera poses, resulting in suboptimal outcomes. To tackle these issues, we propose AltNeRF---a novel framework designed to create resilient NeRF representations using self-supervised monocular depth estimation (SMDE) from monocular videos, without relying on known camera poses. SMDE in AltNeRF masterfully learns depth and pose priors to regulate NeRF training. The depth prior enriches NeRF's capacity for precise scene geometry depiction, while the pose prior provides a robust starting point for subsequent pose refinement. Moreover, we introduce an alternating algorithm that harmoniously melds NeRF outputs into SMDE through a consistence-driven mechanism, thus enhancing the integrity of depth priors. This alternation empowers AltNeRF to progressively refine NeRF representations, yielding the synthesis of realistic novel views. Extensive experiments showcase the compelling capabilities of AltNeRF in generating high-fidelity and robust novel views that closely resemble reality.
\end{abstract}

    \begin{figure}
    \centering
    \includegraphics[width=0.98\linewidth]{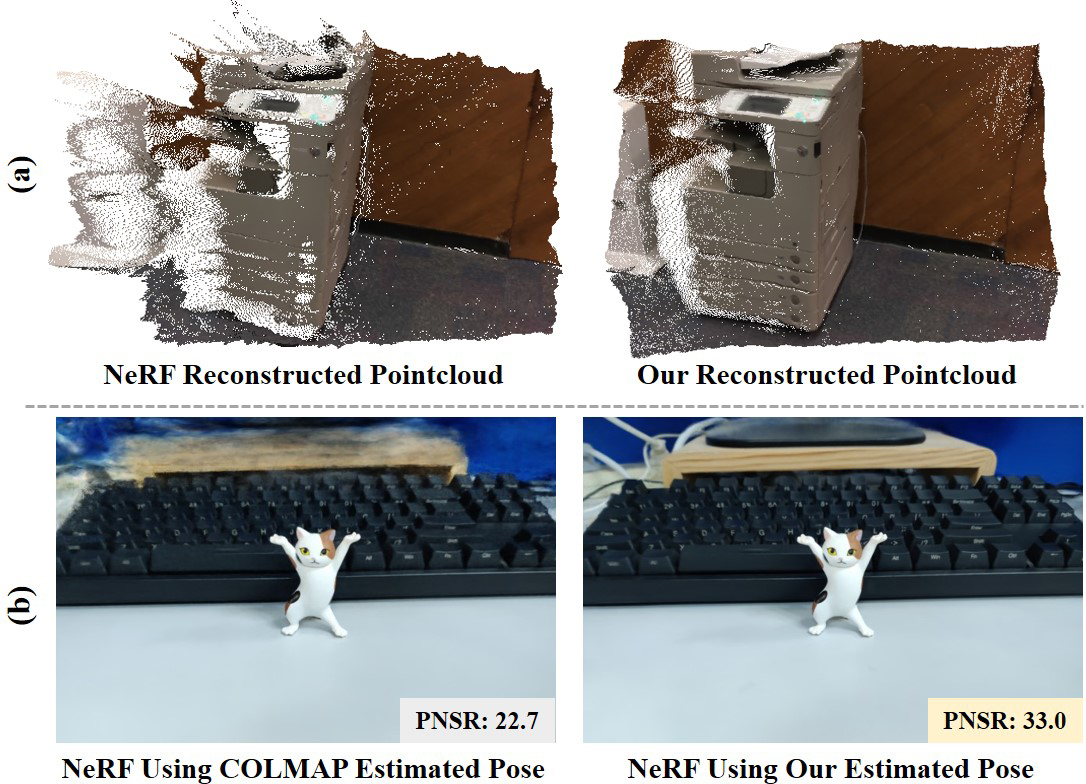}
    \caption{(a) We showcase that NeRF is prone to fitting incorrect geometry, by back-projecting a training view to point cloud using estimated radiance and density. (b) We also show how NeRF creation is affected by inaccurate poses, by optimizing NeRF with COLMAP and our pose estimation.}
    \label{fig.1}
\end{figure}

\section{Introduction}

Neural rendering has achieved unprecedented progress on the long-standing view synthesis task in computer vision research \cite{zhou2016view, kellnhofer2021neural, yu2022anisotropic, li2023read}. A prominent exemplar of this task is NeRF \cite{nerf}, which captures the continuous volumetric essence of real-world scenes using multi-view images and precise camera poses, thereby generating lifelike new perspectives. However, NeRF often struggles with suboptimal outcomes that compromise novel view synthesis and distort scene geometry. In this paper, we identify and address two primary causes for this issue. First, NeRF solely hinges on 2D image supervision, which may provide inadequate geometric constraints for textureless or view-limited scenes. As evidenced in Fig. \ref{fig.1} (a), NeRF is trapped into a suboptimal solution that manifests itself as incorrect scene geometry. Second, NeRF’s reliance on precise camera poses for constructing proper volumetric representations is a stumbling block in the face of pose inaccuracies or noise. Such errors in the camera poses compound the optimization challenges for NeRF, as illustrated in Fig. \ref{fig.1} (b).

\begin{figure}[t]
    \centering
    \includegraphics[width=0.98\linewidth]{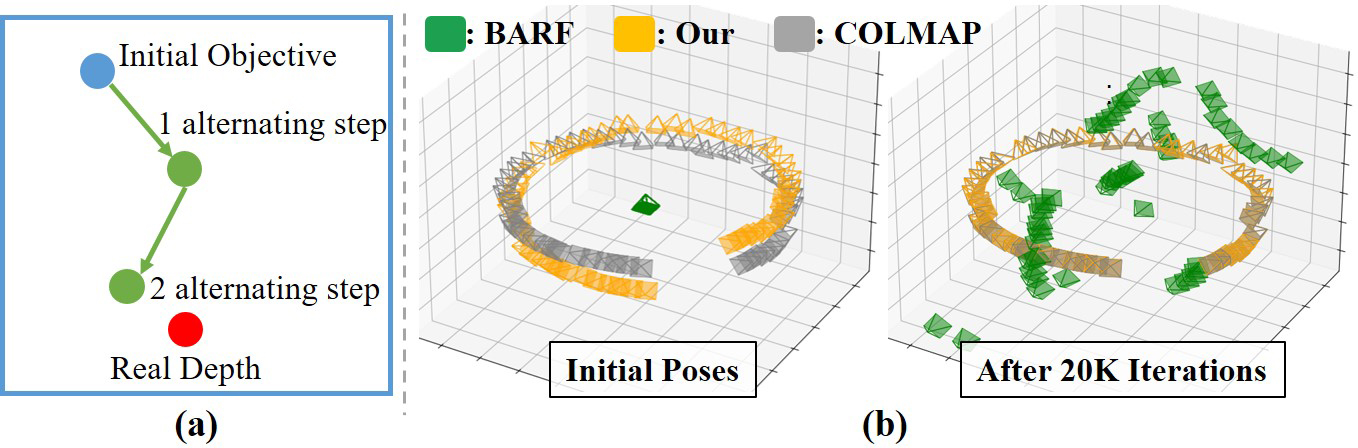}
    \caption{(a) Existing methods establish a fixed target (the blue dot) using inaccurate depth prior, whereas we leverage valuable intermediate results from NeRF to dynamically adjust the objective (the green dots) towards the real depth (the red dot). (b) Pose refinement starting from different initial poses. The experiment is conducted on Vasedeck scene.}
    \label{fig.pose}
\end{figure}

Although existing methods have endeavored to tackle either of these issues, they remain encumbered by certain limitations. Firstly, some methods \cite{ds-nerf, ddp,10.1145/3581783.3612306} leverage depth priors as an explicit 3D supervision to prevent NeRF from fitting incorrect scene geometry. These methods derive depth priors from structure-from-motion (SFM) methodologies or depth estimation techniques, employing them as fixed constraints for NeRF. However, these depth priors might not attain the requisite accuracy, potentially skewing NeRF's optimization trajectory and yielding deteriorated performance. Secondly, some methods \cite{nerfmm,barf,sc-nerf} alternative strategies undertake the joint optimization of NeRF and camera poses to remove the requirement for accurate camera poses. Nevertheless, this combined task encompasses a non-convex optimization conundrum that is acutely sensitive to the initialization of camera poses. Consequently, these approaches necessitate initial camera poses that closely approximate the optimal values; otherwise, they frequently converge towards unfavorable local minima. Illustratively, Fig. \ref{fig.pose} (b) depicts the scenario where BARF \cite{barf} employs identity matrices as green-hued initializations, eventually converging to nonsensical poses after numerous iterations.

To address the previously mentioned problems, we propose AltNeRF—a novel framework designed to generate high-fidelity neural radiance fields from monocular videos. The core concept of AltNeRF is a synergistic process of self-supervised monocular depth estimation (SMDE) and NeRF optimization. SMDE can learn from accessible video data and provide robust depth and pose estimation, which can serve as depth-pose priors for NeRF. Meanwhile, NeRF maintains a continuous volume field, which can represent 3D scenes more accurately than warping-based SMDE. We propose an alternating algorithm that leverages the complementary strengths of SMDE and NeRF to progressively enhance both methodologies. Specifically, we use the pose estimated by SMDE as an effective initialization for NeRF, facilitating smoother optimization akin to the orange poses depicted in Fig. \ref{fig.pose} (b). Furthermore, we use the estimated depth as an initial objective for NeRF, steering it away from reconstructing inaccurate scene geometries. After optimizing NeRF for many iterations, we can obtain improved pose and depth from it, which can further improve the depth estimation of SMDE. This alternation continuously updates the depth objective to converge towards actual scene depths, as illustrated in Fig. \ref{fig.pose} (a). By harnessing the complementary strengths of SMDE and NeRF, AltNeRF can achieve more reliable scene representations. Overall, our contributions can be summarized as: 
\begin{itemize}[leftmargin=*] 
    \item We introduce depth-pose priors learned from monocular videos to simultaneously regularize the scene geometries and initialize the camera poses to enhance the novel view synthesis of NeRF. \item To the best of our knowledge, we are the first to propose AltNeRF—a novel framework that alternately optimizes self-supervised monocular depth estimation and NeRF, synergistically boosting both components. 
    \item We also collect a new dataset of indoor videos captured with a cellphone. Extensive experiments on LLFF, ScanNet, CO3D and our dataset demonstrate that our AltNeRF can synthesize realistic novel views with high fidelity and robustness, and outperforms the existing NeRF methods. 
\end{itemize}

\section{Related Work}

\paragraph{Self-supervised Monocular Depth Estimation.} The learning of SMDE is an image reconstruction problem. It is supervised by the photometric loss that measures the difference between a target frame and frames warped from nearby views. SfM-Learner \cite{sfm-learner} is a seminal work that proposed to jointly predict scene depth and relative camera poses. Follow-up works enhanced SfM-Learner by decomposing depth scale \cite{wang2021can, yan2023desnet}, introducing powerful neural networks \cite{packnet, hr-depth, depth-former}, and applying iterative refinement \cite{dualrefine}. Furthermore, MonoDepth2 \cite{mono2} proposed a minimum reprojection loss to handle occlusions, and some works addressed the dynamic object problem by compensating and masking pixels within dynamic areas using optical flow \cite{df-net, ranjan2019competitive} and pretrained segmentation models \cite{gordon2019depth}. Some other works boosted the performance of self-supervised depth estimation by introducing a feature-metric loss \cite{feature-metric}, proposing a resolution adaptive framework \cite{ra-depth}, and exploring the knowledge distilling approaches \cite{petrovai2022exploiting, ren2022adaptive}. Recently, some works have focused on challenging environments, such as indoor \cite{monoindoor, distdepth} and nighttime \cite{adfa, rnw, liu2021self} scenes and shown impressive performance.

\begin{figure*}
   \centering
   \includegraphics[width=0.78\linewidth]{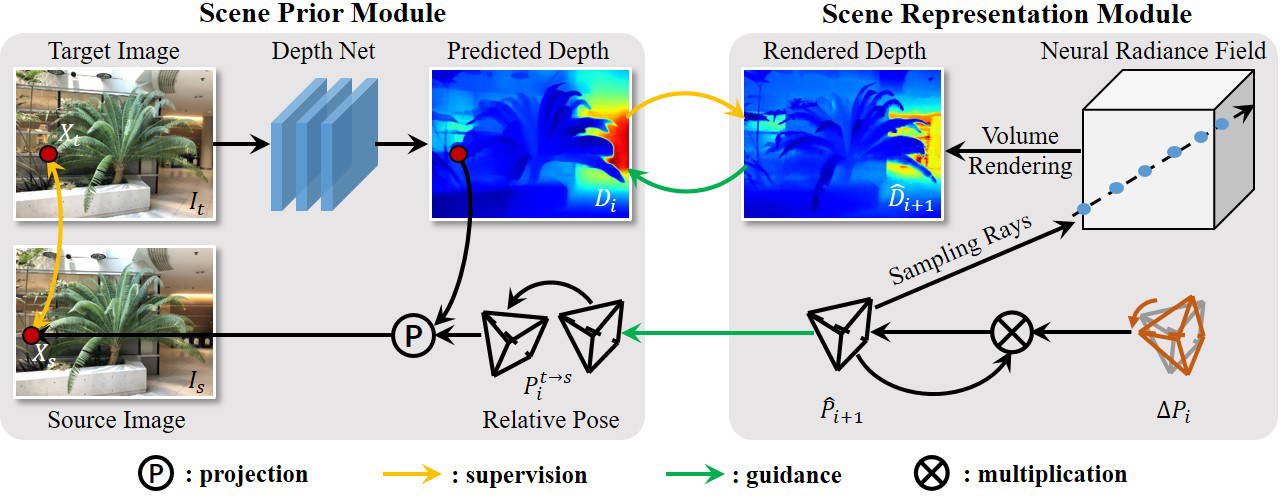}   
   \caption{The overall pipeline of our AltNeRF. The scene prior module estimates depth and pose, which serves as the depth reference and initial poses, respectively. The scene representation module simultaneously refines the initial poses with $\Delta P_i$ and learns 3D scene representation, which is regularized by $D_i$, and produces more accurate poses $\hat{P}_{i+1}$ and finer depth maps $\hat{D}_{i+1}$. These refined depth and pose are then fed back to the scene prior module as guidance to improve its performance. }
   \label{fig.alternating}
\end{figure*}

\paragraph{View Synthesis with NeRF.} NeRF is a powerful technique for novel view synthesis, but they face several challenges in different scenarios. Many works have extended NeRFs to handle dynamic \cite{dnerf,liu2023robust}, unbounded \cite{nerf++, mipnerf360, merf}, and large-scale scenes \cite{block-nerf, meganerf}, as well as to optimize NeRFs from in-the-wild \cite{nerf-wild} and dark images \cite{nerf-dark}. Some works have also improved the generalization \cite{pixelnerf, ibrnet, dbarf}, bundle sampling \cite{adanerf}, initialization \cite{bergman2021fast, tancik2021learned} and data structure \cite{plenoctrees,instantngp} of NeRFs. However, these methods still rely on accurate camera poses, which are not always available or realistic. To address this problem, recent works \cite{nerfmm, sc-nerf, gnerf, barf} have studied the joint task of optimizing NeRF model and camera poses. However, they are restricted to simple or known pose distribution. Moreover, some methods introduce depth priors \cite{ds-nerf, ddp} from external sources, which may be noisy or inaccurate and result in suboptimal NeRF outcomes. In contrast, we introduce SMDE to estimate the depth-pose priors to assist NeRF optimization, and devise an alternating algorithm to harnesses the complementary strengths of SMDE and NeRF for robust NeRF creation.

\section{Preliminary}
In this section, we review the key concepts and techniques of Self-supervised Monocular Depth Estimation (SMDE) and Neural Radiance Field (NeRF) to provide the necessary background for our method.

\paragraph{Self-supervised Monocular Depth Estimation.} SMDE is a training method that only requires monocular videos $\mathcal{V}$ and known camera intrinsic $K$. It employs two neural networks, $f_d: I\rightarrow D$ and $f_p: (I_t, I_s)\rightarrow P_{t\rightarrow s}$, to predict the depth map $D$ of an input image $I$ and relative camera pose $P_{t\rightarrow s}$ between frames $I_t$ and $I_s$. The training objective is to reconstruct the target frame $I_t$ from nearby views $I_s$ by warping pixels $x_s$ from the source image to the target image $x_t$ based on the predicted depth and camera pose: $x_s \sim KP_{t\rightarrow s}D(x_t)K^{-1}x_t$. The photometric loss is used to supervise this process, which consists of the structural similarity term and the $\ell_1$ term:
\begin{equation}
\begin{aligned}
    L_p(I_t,\hat{I_t})=&\frac{\alpha}{2}(1-\mathop{SSIM}(I_t,\hat{I_t}))+\\&(1-\alpha)\|I_t-\hat{I_t}\|_1,
\end{aligned}
\end{equation}
where $\alpha$ is often set to 0.85. An edge-aware smoothness loss is also added to ensure smoothness in predicted depth maps. This loss is based on the image gradients $\partial_x$ and $\partial_y$ along the horizontal and vertical axes, and is weighted by an exponential function of the image gradients to preserve edges:
\begin{equation}\label{eq.Ls}
L_s=|\partial_x D|e^{-|\partial_x I|}+|\partial_y D|e^{-|\partial_y I|},
\end{equation} 
where $\vert\cdot\vert$ returns the absolute value.

\paragraph{Neural Radiance Field.} NeRF represents a scene as a continuous volumetric field. It takes in a 3D point $\mathrm{p}\in\mathbb{R}^3$ and a unit viewing direction $\mathrm{d}\in\mathbb{R}^3$, and returns the corresponding density $\sigma$ and color $c$: $f_n: (\mathrm{p}, \mathrm{d})\rightarrow (\sigma, c)$. The volumetric field can be rendered to 2D images using volume rendering techniques \cite{volumerender}:
\begin{equation}
\hat{C}(r)=\int_{t_n}^{t_f}T(t)\sigma(t)c(t)dt.
\end{equation}
Similarly, the scene depths are created by computing the mean terminating distance of a ray $r=\mathrm{o}+t\mathrm{d}$ parameterized by camera origin $\mathrm{o}$ and viewing direction $\mathrm{d}$, via
\begin{equation}
\hat{D}(r)=\int_{t_n}^{t_f}T(t)\sigma(t)tdt,
\end{equation}
where $T(t)=exp(-\int_{t_n}^t\sigma(s)ds)$ handles occlusions, and $t_n$ and $t_f$ are near and far depth bounds, respectively. The optimization objective of NeRF is to minimize the reconstruction loss, which is computed as the squared differences between the rendered and ground truth colors for all rays:
\begin{equation}
L_c=\Vert\hat{C}(r)-C(r)\Vert_2.
\end{equation}

\section{AltNeRF Framework}
In this section, we introduce our AltNeRF framework, which comprises two components: the Scene Prior Module (SPM) and the Scene Representation Module (SRM). These modules work together under an alternating algorithm. In the following sections, we will delve into the details.

\subsection{Scene Prior Module}

\subsubsection{Pretraining}
NeRF is optimized individually for each scene, and thus lacks the prior knowledge for scene understanding. To address this limitation, we pretrain SPM on a large dataset to accumulate the prior knowledge for 3D recovering. SPM is built on SMDE, which only requires accessible video data for training. However, SMDE is susceptible to dynamic objects and view-dependent appearances, which can degrade its performance. Therefore, we employ the distilling strategy introduced in \cite{distdepth} to mitigate these disadvantages:
\begin{equation}\label{eq.lr}
    \begin{aligned}
        L_r = 1&-\mathop{SSIM}(D, D_r) +\\0.1&\times(E_r \oplus E/\mathop{size}(E)),
    \end{aligned}
\end{equation} 
where $D_r$ is the reference depth map produced by an off-the-shelf relative depth estimator, DPT\cite{dpt}, $\oplus$ denotes XOR operation, $\mathop{size}(\cdot)$ returns the size of a set, and $E_r$ and $E$ are occluding boundary maps of $D_r$ and $D$, respectively. The final loss of this stage is:
\begin{equation}\label{eq.lpt}
    L_{pt}=L_p+L_r+1.0e^{-3}\times L_s.
\end{equation}

\subsubsection{Test-time Adaptation}
The input for AltNeRF is also video data. To adapt SPM to the target video, we fine-tune it for a little iterations using the input video data. This helps close the domain gap between the target and the training data. However, SMDE produces relative depths defined up to an unknown scale factor, which can cause potential inconsistency across frames. Therefore, we introduce the geometry consistency loss from \cite{bian} to ensure the scale consistency:
\begin{equation}
    L_g=\frac{\Vert D_s(x_s)-D_t(x_t)\Vert_1}{D_s(x_s)+D_t(x_t)},
\end{equation}
where $D_s$ and $D_t$ are predicted depth map of $I_s$ and $I_t$, respectively. The final loss used in adaptation step is 
\begin{equation}\label{eq.ad}
    L_{ad} = L_{pt} + 0.5\times L_g.
\end{equation}

\subsubsection{Pose Conversion}
To register each camera to an unified world coordinate, we use the following procedure. We establish a world coordinate system that aligns with the camera coordinate system of the first frame $I_0$, whose pose matrix is an identity matrix. SPM predicts the relative 3D transformations $P_{i-1\rightarrow i}$ between adjacent frames, which we use to calculate the camera poses $P_i$ of subsequent frames. We apply the chain rule to compute the camera poses via $P_{i}=P_{i-1\rightarrow i}\times P_{i-1}$.

\subsection{Scene Representation Module}

SRM serves a dual purpose of learning 3D scene representation and refining camera poses simultaneously. It extends the BARF approach \cite{barf} by introducing three improvements: depth regularization, improved pose initialization, and warmup learning. These enhancements will be discussed in more details below.

\subsubsection{Depth Regularization} 
Recovering 3D geometry from the view-limited scenes (\eg forward-facing scenes) or textureless scenes (\eg indoor scenes) is an ill-posed problem, since there are numerous incorrect shapes that can also explain the input images. To address this problem, we introduce the depth prior estimated by SPM as an explicit 3D supervision for NeRF. This helps mitigate the shape ambiguity that can mislead the NeRF optimization to a degenerate solution. Specifically, we enforce the consistency between depth prior $D$ and NeRF rendered depth $\hat{D}$. However, we do not strictly align the rendered depth with the depth prior as previous works did \cite{ds-nerf, ddp}, since the depth prior itself is not precise either. Instead, we propose an error-tolerant depth regularization that enforces the rendered depth to fit a possible depth range:
\begin{equation}\label{eq.le}
    L_e=H\left(\max\left(\frac{\Vert\hat{D}-D\Vert_1}{\hat{D}+D} - \epsilon, 0\right)\right),
\end{equation}
where $H(\cdot)$ denotes the Huber loss \cite{huber}, and $\epsilon$ is a tolerance coefficient controlling the length of the possible depth range.

\subsubsection{Improved Pose Initialization}
To improve the joint optimization of camera pose and NeRF representation, which is a highly non-convex problem that is prone to converging to a suboptimal solution when the initial pose is far from the actual pose, we use a pose prior estimated by SPM as initialization for SRM. This pose prior, denoted as $P_0$, is closer to the actual pose, which helps it converge to the global minimum. We then refine it by optimizing a residual pose $\Delta P$ that represents the difference between them. The refined pose, denoted by $\hat{P}$, is calculated by combining them through $\hat{P} = \Delta P \times P$.

\subsubsection{Warmup Learning}
To improve the joint optimization of camera pose and scene representation, we propose a warmup learning strategy that synchronizes the learning process for these two tasks. SRM learns the scene representation from scratch, but refines the camera pose using a good pose prior. This asynchrony can result in an incorrect update direction for the camera pose. Therefore, we set the learning rate of $\Delta P$ to a low value at the beginning of the training and linearly increase it to the original learning rate after 1K iterations. This allows SRM to learn a prototype of the scene representation before updating the camera pose.

The final loss for optimizing SRM consists of both reconstruction loss and depth regularization:
\begin{equation}\label{eq.sr}
    L_{sr}=L_c+\gamma\cdot L_e,
\end{equation}
where $\gamma$ is a scalar hyper-parameter that balances these two terms of losses.

\begin{figure}
    \centering
    \includegraphics[width=0.8\linewidth]{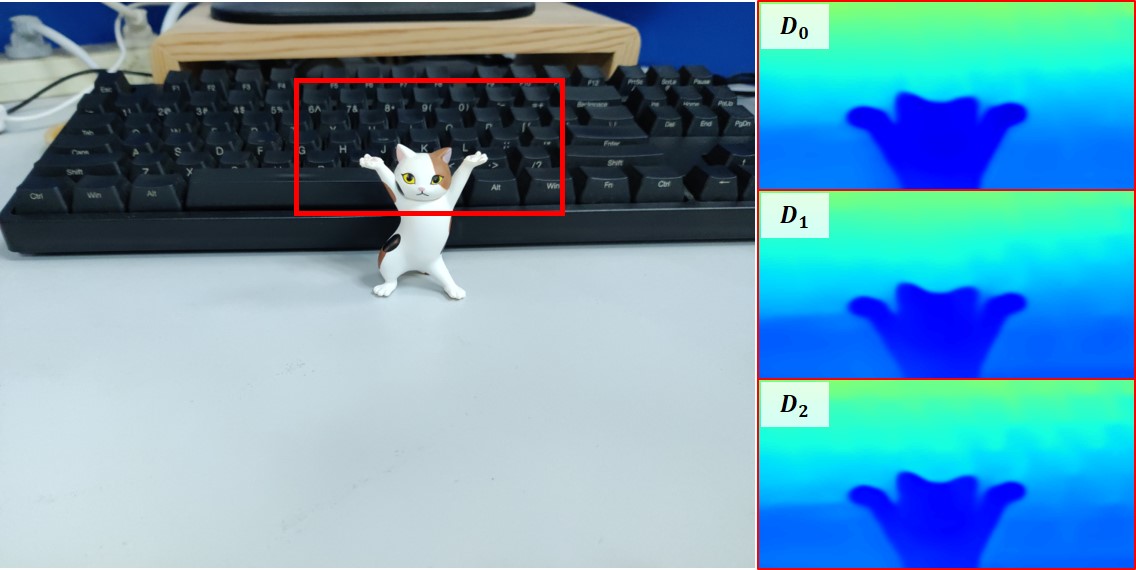} 
    \caption{We illustrate that the depth estimation of SPM is improved by visualizing the initial depth estimate $D_0$, and depth estimates $D_1$ and $D_2$ after 1 and 2 alternating steps.}
    \label{fig.alter_depth}
\end{figure}

\subsection{Alternating Algorithm}

To achieve robust NeRF creation, we propose an alternating algorithm that synergistically boosts SPM and SRM based on their complementary advantages. SPM can produce robust depth and pose estimates as a good scene prior for SRM, but these estimates are not accurate enough to recover precise 3D scenes. On the other hands, SRM learns a continuous volumetric field that can accurately represent 3D space with fine geometries and improves the initial pose through bundle adjustment. Therefore, SRM can overcome the shortcomings of SPM, while an improved SPM can provide more accurate depth regularization which helps SRM achieve better results. The overall framework is illustrated in Fig. \ref{fig.alternating}. In the following sections, we will introduce the workflow of the alternating algorithm, the multi-view consistency check, and discuss its potential extension.

\subsubsection{Workflow}
We denote the alternating step as $i$, SPM at step $i$ as $\Phi_i$, and SRM as $\Psi_i$. The alternation is formulated as:
\begin{equation}
    \begin{aligned}
        &\Psi_i: (D_i, P_i)\rightarrow (\hat{D}_{i+1}, \hat{P}_{i+1}),\\
        &\Phi_i: (\hat{D}_{i+1}, \hat{P}_{i+1})\rightarrow (D_{i+1}, P_{i+1}).
    \end{aligned}
\end{equation} 
In each cycle, SRM takes $P_i$ as initial pose and simultaneously learns scene representation and pose refinement under the regularization of $D_i$. This process is supervised by Eq. \eqref{eq.sr}. After $Sr$ iterations, SRM outputs the estimated scene depth $\hat{D}_{i+1}$ and improved camera pose $\hat{P}_{i+1}$. These improved outcomes are then used to fine-tune the depth estimation network $f_d$ of SPM. Specifically, we use Eq. \eqref{eq.lr} to distill structure information from SRM by replacing the reference depth $D_r$ with rendered depth $\hat{D}_{i+1}$. The final loss that supervises the fine-tuning process is Eq. \eqref{eq.lpt}. We fine-tune SPM for $S_p$ iterations. Through these cyclist steps, both SPM and SRM can be improved. We showcase the improved depth estimation of SPM in Fig. \ref{fig.alter_depth}.

\subsubsection{Multi-view Consistency Check} To assess the possible unreliability of the depth estimates from SPM and SRM, we use a multi-view consistency check to measure the uncertainty of the predicted depth maps. We denote a depth map of a target image $I_t$ as $D_t$, and compute the depth maps $D_{s\rightarrow t}$ warped from nearby source views $I_s$ using camera poses $P_{t\rightarrow s}$ from SPM or $\hat{P}$ from SRM. We expect $D_t$ and $D_{s\rightarrow t}$ to be identical, except for occlusions. Therefore, we define an uncertainty $U_t$ of depth map $D_t$ as the difference between $D_t$ and $D_{s\rightarrow t}$: $U_t = \Vert D_t-D_{s\rightarrow t}\Vert_1$. 
To account for occlusions, we compute the average difference from four views with the smallest differences. We incorporate this depth uncertainty into our loss functions (\ie Eq. \eqref{eq.lr} and Eq. \eqref{eq.le}) by weighting them with the $\text{Softmin}(\cdot)$ function. This helps to mitigate the affect of unreliable depth estimates.

\subsubsection{Discussion} Our alternating algorithm is a general method that can actually leverage any depth-pose priors, not just those learned from SMDE. Using the valuable intermediate results of SPM and SRM, the algorithm can tolerate imprecise priors and still create high-quality NeRF representations, which helps reduce the cost for robust NeRF creation.

\begin{figure*}
    \centering
    \includegraphics[width=0.94\linewidth]{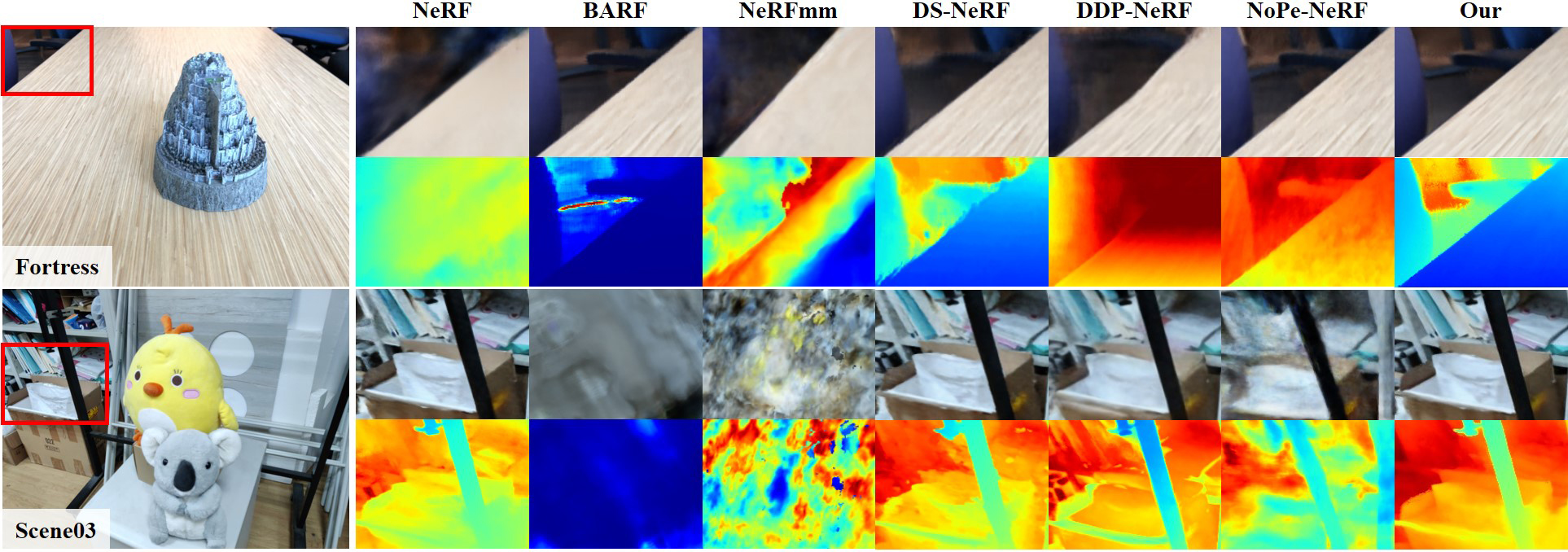}   
    \caption{Qualitative comparisons of novel view synthesis and depth estimation on LLFF and Captures datasets.}
    \label{fig.compare}
\end{figure*}

\section{Experiment}

In this section, we evaluate AltNeRF on 16 scenes of four datasets and compare it with existing methods to demonstrate its state-of-the-art (SOTA) performance. We first introduce the datasets and implementation details, and then report the experiment results.

\subsection{Dataset} We evaluate AltNeRF on four datasets: LLFF \cite{llff}, CO3D \cite{co3d}, ScanNet \cite{scannet} and our collected dataset, named Captures. Each dataset contains different scenes with varying levels of complexity and camera motion. We employ the same train/test data division as BARF, which uses the first 90\% of frames for training and the remaining 10\% for testing. \textbf{LLFF}: we select five scenes from LLFF for evaluation: Fern, Flower, Fortress, Orchids and Room. We also incorporate Vasedeck from nerf\_llff\_360 into this set. \textbf{CO3D}: we randomly select three scenes from the Couch category of this dataset: \emph{193\_20797\_40499}, \emph{349\_36504\_68102} and \emph{415\_57184\_110444}. These three scenes have more than 80 frames per scene and exhibit complex camera motions with simultaneous panning and rotation. \textbf{ScanNet}: we randomly select three scenes, \emph{scene0079\_00}, \emph{scene0553\_00} and \emph{scene0653\_00}, to evaluate the depth estimation performance of AltNeRF. We use the data processed by NerfingMVS \cite{nerfingmvs} and reduce each scene to 20 frames to enlarge the viewpoint difference of adjacent frames. \textbf{Captures}: Captures consists of four scenes that we collect with our smartphone. Two scenes are forward-facing (Scene\_01 and Scene\_02) and two scenes are inner-facing with semicircular camera trajectory (Scene\_03 and Scene\_04). This dataset is challenging as it contains many textureless views and its viewpoint diversity is also limited.

\subsection{Implementation Detail}

The depth estimation network $f_d(\cdot)$ in SPM is based on the U-Net \cite{u-net} architecture. The encoder is a ResNet-50 \cite{resnet} with the fully-connected layer removed, and the decoder consists of ten $3\times 3$ convolutional layers, two for each scale, and uses bilinear up-sampling. The pose estimation network $f_p(\cdot, \cdot)$ is structured with a ResNet-34 and outputs a vector of nine element length, where the first six elements are continuous rotation representation \cite{continuousR} and the last three elements denote translations. SPM is pretrained on around 300K images collected from NYUv2 \cite{nyu} and VOID \cite{void}. The scene representation function $f_n(\cdot, \cdot)$ in SRM uses the same network structure as NeRF, \ie eight fully-connected layers with skip connections for density output, and one linear layer for color.

The $\gamma$ in Eq. \eqref{eq.sr} is set to $0.08$ for LLFF and CO3D, and $0.15$ for ScanNet and Captures. We pretrain the SPM with a learning rate of $10^{-4}$, and fine-tune it with $5.0\times 10^{-5}$. For SRM, the initial learning rate for NeRF learning is set to $10^{-3}$, and exponentially decays to $10^{-4}$ throughout the training process. The initial learning rate for pose refinement is set to $10^{-5}$, and linearly increases to $2.0\times 10^{-3}$ after 1K iterations before exponentially decaying to $10^{-5}$. The number of iterations $S_r$ and $S_p$ are set to 50K and $500$, respectively, and we perform two alternations in all experiments unless otherwise specified. Our method is trained for 150K-200K iterations according to the number of frames, which costs around 4.0-6.4 hours totally on single RTX 3090.

\begin{table}
    \centering
    \setlength{\tabcolsep}{1.8pt}
    \renewcommand\arraystretch{1.1}
    \resizebox{\linewidth}{!}{
        \begin{tabular}{c||ccc|ccc}
            \hline
            & \multicolumn{3}{c|}{\emph{LLFF}}                              & \multicolumn{3}{c}{\emph{Captures}}                           \\ 
            \multirow{-2}{*}{Method} & PSNR$\uparrow$ & SSIM$\uparrow$ &LPIPS $\downarrow$ & PSNR$\uparrow$ & SSIM$\uparrow$ & LPIPS$\downarrow$ \\ \hline
            NeRF                    & {\underline {26.36}}    & {\underline{0.778}}     & {\underline{0.147}}        & 25.48          & 0.886           & {\underline{0.098}}        \\
            BARF            & 24.80          & 0.722           & 0.262              & 19.33          & 0.705           & 0.425              \\
            NeRFmm       & 20.40          & 0.557           & 0.451              & 19.61          & 0.670           & 0.402              \\
            DS-NeRF     & 26.11          & 0.773           & 0.174              & {\underline{27.35}}    & {\underline{0.893}}     & 0.109              \\
            DDP-NeRF     & 22.33          & 0.690           & 0.242              & 24.20          & 0.848           & 0.148              \\
            NoPe-NeRF    & 24.77          & 0.714           & 0.265              & 23.50          & 0.786           & 0.232              \\ \hline
            \rowcolor[HTML]{EFEFEF} 
            Our                      & \textbf{27.41} & \textbf{0.803}  & \textbf{0.139}     & \textbf{29.72} & \textbf{0.922}  & \textbf{0.067}     \\ \hline
        \end{tabular}
    }
    \caption{Quantitative comparison on novel view synthesis. The best result is in bold, and the second is underlined.}
    \label{tab.compare}
\end{table}

\subsection{Comparing with Existing Method}
Here, we evaluate AltNeRF on novel view synthesis, depth estimation and camera pose estimation tasks, and compare it with existing methods to showcase its SOTA performance.

\begin{table}
    \centering
    \setlength{\tabcolsep}{2.8pt}
    \renewcommand\arraystretch{1.1}
    \resizebox{\linewidth}{!}{
        \begin{tabular}{c||cc|cc|cc} 
            \hline
            \multirow{2}{*}{Method}               & \multicolumn{2}{c|}{\emph{Fortress}}    & \multicolumn{2}{c|}{\emph{Orchids}}     & \multicolumn{2}{c}{\emph{Vasedeck}}      \\
            & PSNR$\uparrow$        & SSIM$\uparrow$          & PSNR$\uparrow$           & SSIM$\uparrow$          & PSNR$\uparrow$            & SSIM$\uparrow$           \\ 
            \hline
            DS-BARF                               & 30.61          & 0.883          & 19.46          & 0.582          & 21.83          & 0.641           \\
            \rowcolor[HTML]{EFEFEF} Our & \textbf{30.98} & \textbf{0.899} & \textbf{20.33} & \textbf{0.648} & \textbf{23.34} & \textbf{0.701}  \\
            \hline
        \end{tabular}
    }
    \caption{Quantitative comparison on novel view synthesis with COLMAP assisted baseline, DS-BARF.}
    \label{tab.ds-barf}
\end{table}

\subsubsection{Evaluation on LLFF and Captures}

We compare AltNeRF with existing methods on novel view synthesis task. The compared methods are NeRF \cite{nerf}, BARF \cite{barf}, NeRFmm \cite{nerfmm}, DS-NeRF \cite{ds-nerf}, DDP-NeRF \cite{ddp} and NoPe-NeRF \cite{nopenerf}. These methods aim to address the shape ambiguity or camera pose requirement of NeRF. Tab. \ref{tab.compare} shows the mean quantitative results. We use PSNR, SSIM \cite{ssim} and LPIPS \cite{lpips} metrics to evaluate the image synthesis performance. We employ identity matrices to initialize the camera pose of BARF and NeRFmm. The LLFF and Captures datasets do not have dense depth ground truth, thus we directly use the pretrained depth completion model provided by DDP-NeRF, as its dense depth prior. In general, our method outperforms the competitors on all metrics. On LLFF, it outperforms the second best method, NeRF, by 3.98\%, 3.21\% and 5.44\% on PSNR, SSIM and LPIPS, respectively. The improvement on Captures is more significant since the competitors perform pool on textureless and view-limited scenes. Specifically, AltNeRF improves the second best method, DS-NeRF, by 8.66\%, 3.25\% and 38.53\% on these three metrics.

\begin{table}
    \centering
    \setlength{\tabcolsep}{3.5pt}
    \renewcommand\arraystretch{1.1}
    \resizebox{\linewidth}{!}{
        \begin{tabular}{c||ccc|ccc}
            \hline
            Method     & Abs Rel$\downarrow$ & Sq Rel$\downarrow$ & RMSE$\downarrow$ & $\sigma_1$$\uparrow$ & $\sigma_2$$\uparrow$ & $\sigma_3$$\uparrow$ \\ \hline
            NeRF       & 0.143                & 0.072               & 0.312             & 80.5                 & 95.8                 & 96.7                 \\
            DS-NeRF    & {\underline{0.075}}          & {\underline{0.025}}         & 0.169             & 90.4                 & 95.6                 & 99.5                 \\
            NerfingMVS & {\underline{0.075}}          & {\underline{0.025}}         & {\underline{0.164}}       & {\underline{93.8}}           & {\underline{98.9}}           & {\underline{99.8}}           \\ \hline
            \rowcolor[HTML]{EFEFEF} 
            Our        & \textbf{0.051}       & \textbf{0.008}      & \textbf{0.106}    & \textbf{98.7}        & \textbf{99.8}        & \textbf{99.9}        \\ \hline
        \end{tabular}
    }
    \caption{Quantitative results of depth estimation. The reported results are average over three scenes of ScanNet.}
    \label{tab.sn}
\end{table}

\begin{table}
    \centering
    \setlength{\tabcolsep}{2.25pt}
    \renewcommand\arraystretch{1.1}
    \resizebox{\linewidth}{!}{
        \begin{tabular}{c||cc|cc|cc}
            \hline
            & \multicolumn{2}{c|}{\emph{Couch\_193}} & \multicolumn{2}{c|}{\emph{Couch\_349}} & \multicolumn{2}{c}{\emph{Couch\_415}} \\
            \multirow{-2}{*}{Method} & PSNR$\uparrow$   & SSIM$\uparrow$  & PSNR$\uparrow$   & SSIM$\uparrow$  & PSNR$\uparrow$   & SSIM$\uparrow$  \\ \hline
            NeRF              & {\underline{ 34.56}}      & 0.929      & {\underline{32.46}}      & {\underline{0.906}}      & {\underline{36.07}}      & {\underline{0.948}}      \\
            BARF         & 9.42             & 0.384            & 12.89            & 0.547            & 15.78            & 0.717            \\
            DS-NeRF       & 32.48            & \textbf{0.939}            & 31.19            & 0.887            & 34.97            & 0.928            \\
            NoPe-NeRF   & 15.21            & 0.495            & 13.09            & 0.593            & 16.35            & 0.698            \\ \hline
            \rowcolor[HTML]{EFEFEF} 
            Our                      & \textbf{34.69}   & \underline{0.930}   & \textbf{33.08}   & \textbf{0.912}   & \textbf{37.08}   & \textbf{0.949}   \\ \hline
        \end{tabular}
    }
    \caption{Quantitative comparison on novel view synthesis task. The experiments are conducted on CO3D.}
    \label{tab.co3d}
\end{table}

Existing methods focus on either addressing the shape ambiguity or the pose requirement of NeRF, while our method addresses these two problems simultaneously. To further demonstrate the advantages of our method, we combine BARF with DS-NeRF as a new baseline, which can also optimize pose and regularize NeRF learning. We denote this new baseline as \emph{DS-BARF}. It initializes the camera pose using COLMAP \cite{schoenberger2016sfm} estimated pose, and supervises NeRF learning using the same method as DS-NeRF, with the depth estimated by COLMAP. We freeze its camera pose for the first 1K iterations to align the learning process of camera pose and NeRF. We compare with this baseline on three scenes of LLFF, and report the quantitative results in Tab. \ref{tab.ds-barf}. The results demonstrates that simply combining existing methods is insufficient for high-quality NeRF creation.

Fig. \ref{fig.compare} shows the qualitative comparisons on novel view synthesis and depth estimation tasks. Each method is evaluated on two scenes from LLFF and Captures datasets. The results show that AltNeRF can synthesize realistic novel views and reasonable depth maps for both scenes. For example, it estimates the depth of the distant chairs in the Fortress scene more accurately, while the other methods underestimate their depth or fail to capture their details.

\subsubsection{Evaluation on ScanNet}

\begin{figure}
    \centering
    \includegraphics[width=0.96\linewidth]{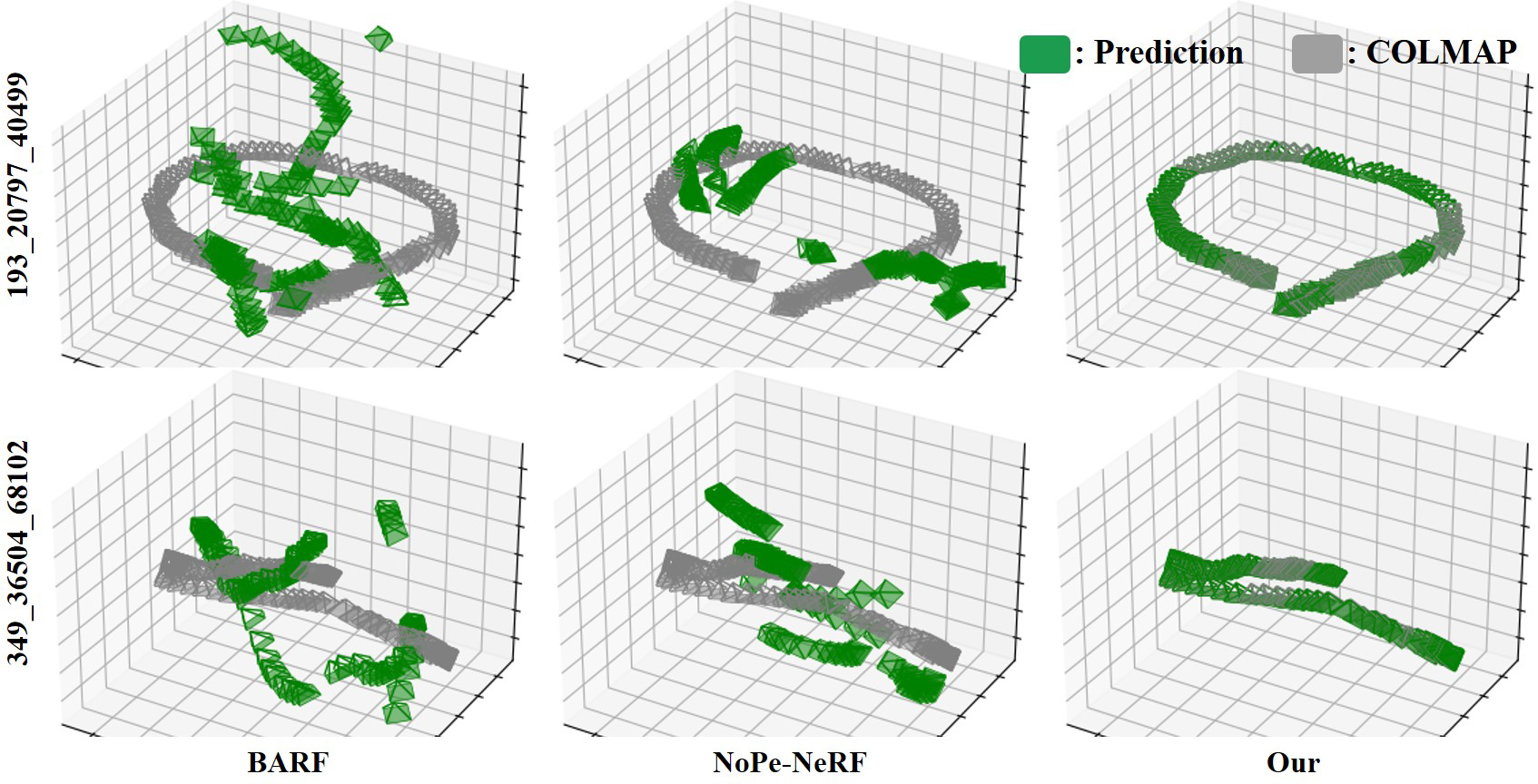}
    \caption{Qualitative results of pose estimation on CO3D.}
    \label{fig.pose_vis}
\end{figure}

\begin{table}
    \centering
    \setlength{\tabcolsep}{1.5pt}
    \renewcommand\arraystretch{1.1}
    \resizebox{\linewidth}{!}{
        \begin{tabular}{l||ccc|ccc} 
            \hline
            \multirow{2}{*}{Method}  & \multicolumn{3}{c|}{\emph{Flower}}                           & \multicolumn{3}{c}{\emph{Scene\_02}}                           \\
            & PSNR$\uparrow$ & SSIM$\uparrow$ & LPIPS$\downarrow$ & PSNR$\uparrow$ & SSIM$\uparrow$ & LPIPS$\downarrow$  \\ 
            \hline
            BARF                     & 23.99          & 0.744           & 0.183              & 29.68          & 0.932           & 0.053               \\
            + pose prior             & 25.21          & 0.752           & 0.154              & 31.31          & 0.962           & 0.038               \\
            + depth prior            & 24.99          & 0.757           & 0.141              & 34.09          & 0.974           & 0.030                \\ 
            \hline
            1 alternation  & 25.96          & 0.786           & 0.117              & 35.05          & 0.976           & 0.030                \\
            2 alternations  & \underline{26.07}  & \textbf{0.794}  & \underline{0.114}      & \underline{35.05}  & \underline{0.978}   & \underline{0.029}       \\
            4 alternations & \textbf{26.13} & \underline{0.793}   & \textbf{0.112}     & \textbf{35.05} & \textbf{0.978}  & \textbf{0.028}      \\
            \hline
        \end{tabular}
    }
    \caption{Quantitative results for ablation study. BARF is the baseline method and we gradually enable each component to demonstrate their effectiveness.}
    \label{tab.ablation}
\end{table}

ScanNet has high-quality depth ground truth, therefore we use it to demonstrate the superior geometry reconstruction ability of AltNeRF. Specifically, we compare AltNeRF with three existing methods, namely NeRF, DS-NeRF, and NerfingMVS, and report the quantitative results in Tab. \ref{tab.sn}. We use three error metrics, Abs Rel, Sq Rel and RMSE, and three accuracy metrics, $\sigma_1$, $\sigma_2$ and $\sigma_3$ (\%), to measure the quality of the estimated depth maps. Our method outperforms the competitors by a large margin on each scene. It reduces the Sq Rel and RMSE metrics by 68.0\% and 35.37\%, respectively, compared to the second best method, NerfingMVS. It also achieves a near-perfect performance on the $\sigma_3$ metric, which indicates a high accuracy of depth estimation. This demonstrates that AltNeRF can perform well on the depth estimation task, and also shows that it can learn a more accurate scene representation than the existing methods.

\subsubsection{Evaluation on CO3D}

CO3D dataset contains long-length videos with more complicated room-level camera motion. We use this dataset to demonstrate that AltNeRF is also applicable in these challenging scenes. First, we report the quantitative results on novel view synthesis task in Tab. \ref{tab.co3d}. Note that, SFM methods (\eg COLMAP) usually work well when input images are abundant. However, AltNeRF still outperforms NeRF and DS-NeRF, which employ the camera pose and depth estimated by COLMAP. This demonstrates the superiority of our method over the COLMAP assisted approaches. To evaluate AltNeRF on the camera pose estimation task, we use the camera pose estimated by COLMAP as pseudo ground truth and report the qualitative comparison with BARF and NoPe-NeRF in Fig. \ref{fig.pose_vis}. Our predictions are closely coincide with those of COLMAP, while BARF and NoPe-NeRF fail to produce meaningful pose outputs.

\subsection{Ablation Study}

\begin{figure}
    \centering
    \includegraphics[width=0.96\linewidth]{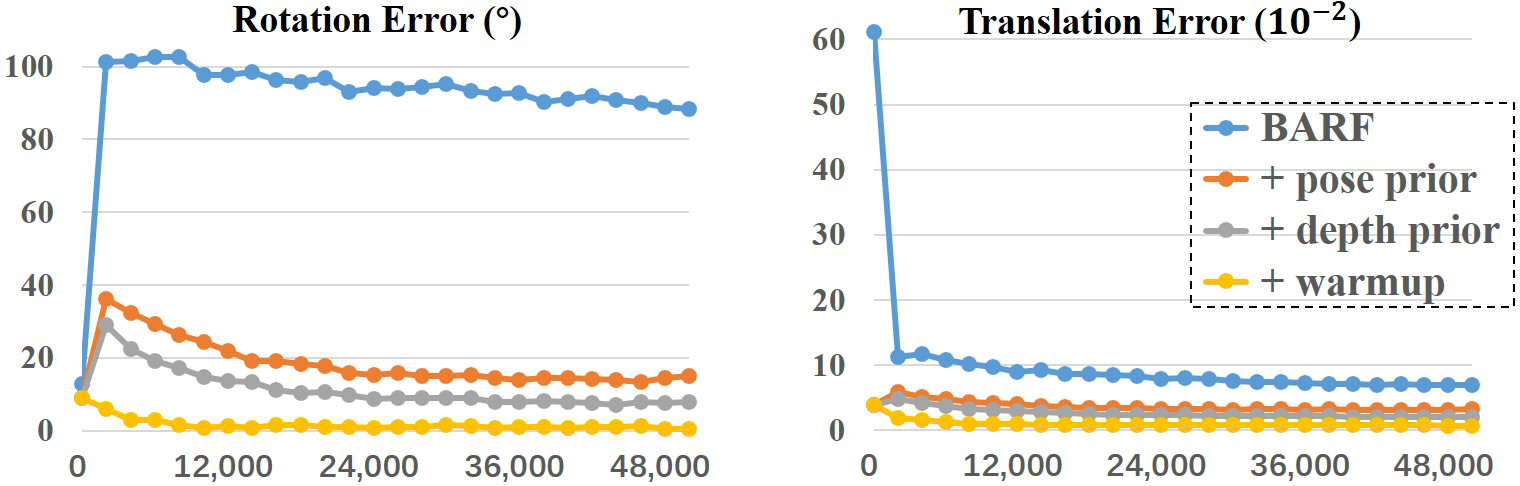}
    \caption{Ablation study on pose estimation on Scene\_04. The x-axis indicates the number of iterations. BARF is the baseline method, and we gradually enable each component to test their performance.}
    \label{fig.ablation_pose}
\end{figure}

We demonstrate the effectiveness of each component of AltNeRF through ablation study reported in Tab. \ref{tab.ablation}. We use BARF with identity matrices initialization as the baseline. First, we introduce the pose prior by initializing the camera poses with SPM estimates. This improves the performance on all metrics. Then, we introduce the depth prior and regularize NeRF with the proposed error-tolerant loss. This improves most metrics except for PSNR of Flower. We attribute this to the inaccurate depth prior, which impairs the learning of scene representation. The inaccurate depth prior can be improved via alternation. We report the results using 1, 2 and 4 alternations in the last three rows of this table. More alternating steps can consistently improve the performance, although the improvement is decreasing.

We evaluate the performance of each component on pose estimation and report the results in Fig. \ref{fig.ablation_pose}. We use BARF as baseline and gradually enable the pose priors, the error-tolerant depth loss and the warmup learning strategy to test their effects. The results show that the camera pose errors decrease as more components are enabled. In particular, enabling the pose priors significantly reduces the pose error. The error-tolerant loss also improves the performance over \emph{+ pose prior}, which verifies its effectiveness. With the warmup learning strategy, the errors are further reduced, leading to the most accurate pose estimation.

\section{Conclusion}
NeRf creation often suffers from suboptimal solutions due to the lack of explicit 3D supervision and imprecise camera poses. In this paper, we propose a alternation-based framework that harmoniously melds self-supervised depth estimation and neural rendering to address these problems. Our method can produce high-quality NeRF representations and accurate camera poses only from monocular videos.

\section{Acknowledgments}
The authors would like to thank the editor and the anonymous reviewers for their critical and constructive comments and suggestions. This work was partially supported by the National Natural Science Foundation of China under Grant 62361166670, 62072242 and 62376121, the Fundamental Research Funds for the Central Universities under Grant 070-63233084, the Young Scientists Fund of the National Natural Science Foundation of China under Grant 62206134 and the Tianjin Key Laboratory of Visual Computing and Intelligent Perception. Note that the PCA Lab is associated with Key Lab of Intelligent Perception and Systems for High-Dimensional Information of Ministry of Education, and Jiangsu Key Lab of Image and Video Understanding for Social Security, School of Computer Science and Engineering, Nanjing University of Science and Technology.

\bibliography{references}

\end{document}